\documentclass[10pt,twocolumn,letterpaper]{article}

\usepackage{cvpr}              % To produce the CAMERA-READY version
\usepackage{graphicx}
\usepackage{amsmath}
\usepackage{amssymb}
\usepackage{booktabs}
\usepackage{commath}
\usepackage{amsmath,stackengine}
\stackMath

\usepackage[table]{xcolor}
\usepackage{caption} 
\usepackage[pagebackref,breaklinks,colorlinks]{hyperref}

\usepackage[capitalize]{cleveref}
\crefname{section}{Sec.}{Secs.}
\Crefname{section}{Section}{Sections}
\Crefname{table}{Table}{Tables}
\crefname{table}{Tab.}{Tabs.}

\def\cvprPaperID{9962} % *** Enter the CVPR Paper ID here
\def\confName{CVPR}
\def\confYear{2022}

\begin{document}

\title{AdaptPose: Cross-Dataset Adaptation for 3D Human Pose Estimation by Learnable Motion Generation}

\author{Mohsen Gholami, Bastian Wandt, Helge Rhodin,  Rabab Ward, and Z. Jane Wang \\
University of British Columbia\\
{\tt\small \{mgholami, rababw, zjanew\}@ece.ubc.ca}    
{\tt\small  \{wandt
 rhodin\}@cs.ubc.ca}}

% \author{Mohsen Gholami\\
% University of British Columbia\\
% {\tt\small mgholami@ece.ubc.ca}
% \and
\maketitle

%%%%%%%%% ABSTRACT
\begin{abstract}

This paper addresses the problem of cross-dataset generalization of 3D human pose estimation models. 
Testing a pre-trained 3D pose estimator on a new dataset results in a major performance drop.
Previous methods have mainly addressed this problem by improving the diversity of the training data. 
We argue that diversity alone is not sufficient and that the characteristics of the training data need to be adapted to those of the new dataset such as camera viewpoint, position, human actions, and body size.  
To this end, we propose \emph{AdaptPose}, an end-to-end framework that generates synthetic 3D human motions from a source dataset and uses them to fine-tune a 3D pose estimator.
AdaptPose follows an adversarial training scheme. 
From a source 3D pose the generator generates a sequence of 3D poses and a camera orientation that is used to project the generated poses to a novel view. 
Without any 3D labels or camera information AdaptPose successfully learns to create synthetic 3D poses from the target dataset while only being trained on  2D poses. 
In experiments on the Human3.6M, MPI-INF-3DHP, 3DPW, and Ski-Pose datasets our method outperforms previous work in cross-dataset evaluations by 14\% and previous semi-supervised learning methods that use partial 3D annotations by 16\%.
\end{abstract}

\begin{figure}[t]
\begin{center}

\includegraphics[scale=0.38]{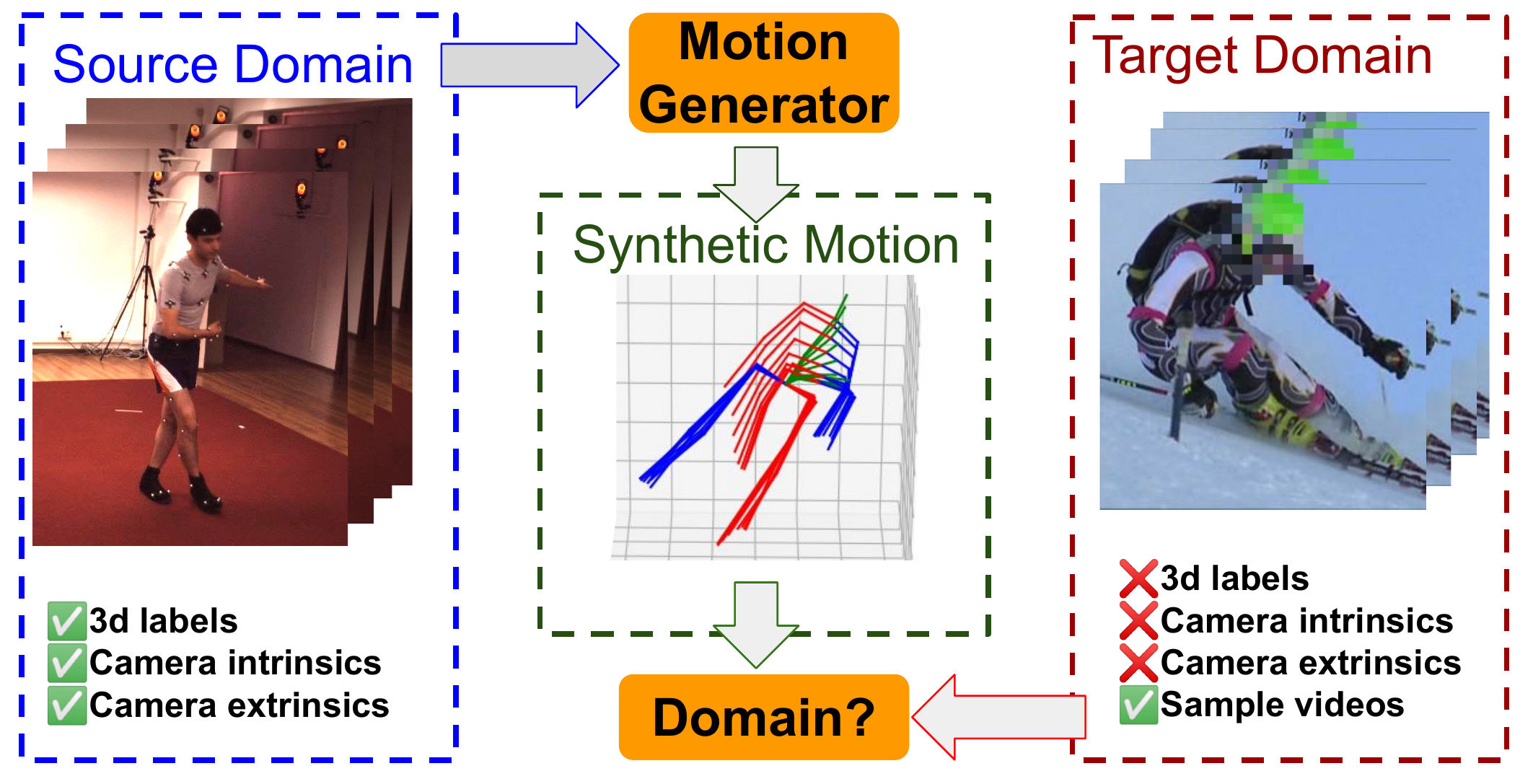}
\end{center}
  \caption{AdaptPose generates synthetic motions to improve the cross-dataset generalization. The source dataset has 3D labels and camera information, while the target dataset has only sample videos.  The synthetic motions are generated  to belong to the target dataset. Therefore fine-tuning the 3D pose estimator with synthetic motions improves the generalization of the model.}
\label{fig:tiser}
\end{figure}
%%%%%%%%% BODY TEXT
\section{Introduction}
\label{sec:intro}

Monocular 3D human pose estimation aims to reconstruct the 3D skeleton of the human body from 2D images. 
Due to pose and depth ambiguities, it is well known to be an inherently ill-posed problem.
However, deep learning models are able to learn 2D to 3D correspondences and achieve impressively accurate results when trained and tested on similar datasets \cite{Martinez_2017_ICCV, pavllo:videopose3D:2019,Arnab_2019_CVPR,Cheng_Yang_Wang_Tan_2020,Hossain_2018_ECCV,pavlakos2018learning,Cai_2019_ICCV}.

An often disregarded aspect is that the distribution of features in a dataset e.g. camera orientation and body poses differ from one dataset to another.
Therefore, a pre-trained network underperforms when applied to images captured from a different viewpoint or when they contain an activity that is not present in the training dataset \cite{Viewpoint_ECCV2020,SRnet_ECCV2020}. 
As an example, Figure \ref{fig:tiser} shows images from the Human3.6M ~\cite{h36m_pami} dataset on the left and images from the Ski-Pose \cite{skipose,Rhodin_2018_CVPR} dataset on the right which we define as \textit{source domain} and \textit{target domain}, respectively. 
Camera viewpoint, position, human action, speed of motion, and body size significantly differ between the source and target domain. This large domain gap causes 3D pose estimation models trained on the source domain to make unreliable predictions for the target domain \cite{Viewpoint_ECCV2020, zhang2020inference,SRnet_ECCV2020}. 
We address this problem by generating synthetic 3D data that lies within the distribution of the target domain and fine-tuning the pose estimation network by the generated synthetic data. Our method does not require 3D labels or camera information from the target domain and is only trained on sample videos from the target domain.

To the best of our knowledge, there are only two approaches that generate synthetic 2D-3D human poses for cross-dataset generalization of 3D human pose estimators ~\cite{Li_2020_CVPR, Guan_2021_CVPR}. Li \etal \cite{Li_2020_CVPR} randomly generate new 2D-3D pairs of the source dataset by substituting parts of the human body in 3D space and projecting the new 3D pose to 2D. PoseAug \cite{Gong_2021_CVPR} proposes a differential data augmentation framework that is trained along with a pose estimator. Both, \cite{Li_2020_CVPR} and \cite{Gong_2021_CVPR}, merely improve the diversity of the source domain without considering the distribution of the target domain. Moreover, these methods are based on single images and do not consider temporal information.

\begin{figure*}[t]
\begin{center}
% \fbox{\rule{0pt}{2in} \rule{.4\linewidth}{0pt}}
\includegraphics[scale=0.6]{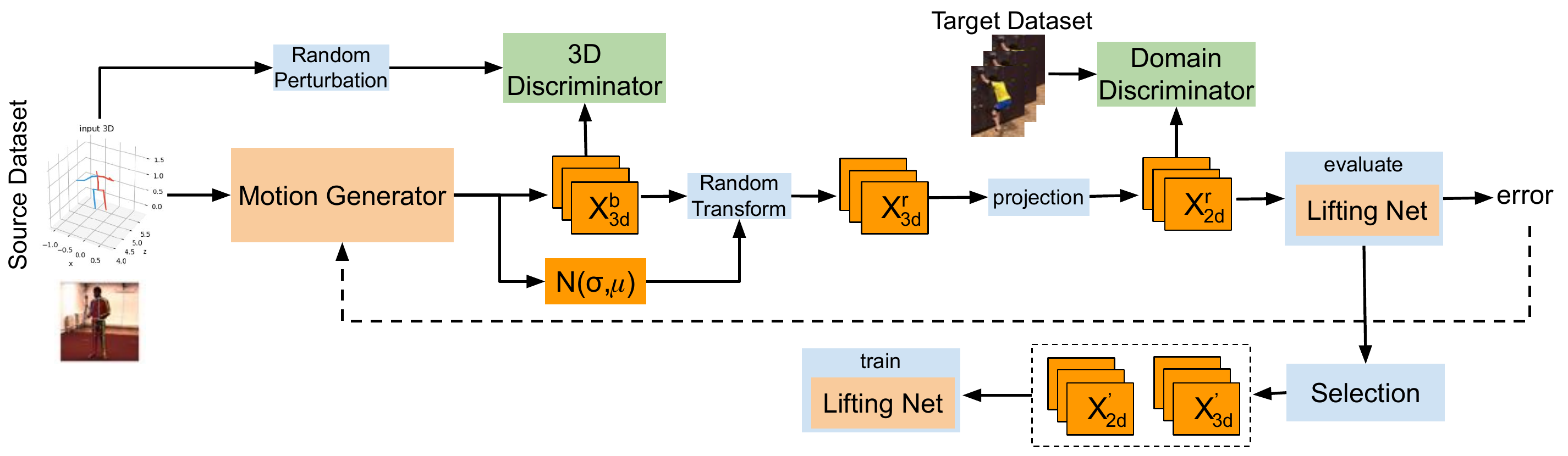}
\end{center}
  \caption{Overview of the proposed network. The input is a vector of 3D keypoints from the source dataset concatenated with Gaussian noise. The motion generator learns to generate a sequence of 3D keypoints $X_{3D}^b$ and the mean and standard deviation of a normal distribution $\mathcal{N}$. A random rotation matrix is sampled from the learned normal distribution and $X_{3D}^b$ is transformed to $X_{3D}^r$ and projected to 2D. The domain discriminator is trained with $X_{2D}^r$ and 2D keypoints from the target domain. The lifting network is a pretrained pose estimator that estimates 3D from 2D. It is used to evaluate $X_{2D}^r,X_{3D}^r$, provide feedback to the motion generator, and to select a subset of samples for fine-tuning the lifting network. The pipeline is trained end-to-end.}
\label{fig:overview}
\end{figure*}

We formulate the data augmentation process as a domain adaptation problem. Figure \ref{fig:overview} shows our training pipeline. Our goal is to generate plausible synthetic 2D-3D pairs that lie within the distribution of the target domain. Our framework, \emph{AdaptPose}, introduces a human motion generator network that takes 3D samples from the source dataset and modifies them by a learned deformation to generate
a sequence of new 3D samples. We project the generated 3D samples to 2D and feed them to a domain discriminator network. The domain discriminator is trained with real 2D samples from the target dataset and fake samples from the generator. We use the generated samples to fine-tune a pose estimation network. Therefore, our network adapts to any target using only images from the target dataset. 3D annotation from the target domain is not required. Unlike \cite{Li_2020_CVPR,Guan_2021_CVPR}, this enables our network to generate plausible 3D poses from the target domain. 
Another contribution is the extension of the camera viewpoint generation from a deterministic approach to a probabilistic approach. We assume that the camera viewpoint of the target domain comes from a specific well-defined, but unknown distribution. Therefore, we propose to learn a distribution of camera viewpoints instead of learning to generate a deterministic rotation matrix. 
Our network rotates the generated 3D poses into a random camera coordinate system within the learned distribution. The generated sample is a sequence of 2D-3D pose  pairs that entails plausibility in the temporal domain. We believe that the application of the proposed motion generator is not limited to  improving only cross-dataset performance of 3D pose estimation, but it could also be used in other tasks such as human action recognition.

\textbf{Contributions.} 1) we propose to close the domain gap between the training and test datasets by a kinematics-aware domain discriminator. The domain discriminator is trained along with a \textit{human motion generator} (HMG) that  uses a source training dataset to generate human motions close to those in the target dataset. 2) We show that learning the distribution of the camera viewpoint is more effective than learning to generate a deterministic camera matrix. 3) To the best of our knowledge, this is the first approach that proposes generating human motions specifically for cross-dataset generalization for 3D human pose estimation, unlike previous work that focuses on single-frame data augmentation.

\section{Related Work}
\label{sec:relatedwork}
In the following, we discuss the related work with a focus on cross-dataset adaptation.

\textbf{Weakly-supervised Learning.} Weakly supervised learning has been proposed to diminish the dependency of networks on 3D annotations. These methods rely on unpaired 3D annotation \cite{Wandt_2019_CVPR,Kundu_2020_CVPR,Yang_2018_CVPR}, multi-view images \cite{gholami2021tripose,Iqbal_2020_CVPR,Wandt_2021_CVPR,Rhodin_2018_CVPR,Kocabas_2019_CVPR}, or cycle-consistency \cite{Chen_2019_CVPR, Drover_2018_ECCV_Workshops}. Most related to our work is the adaptation of a network to the target domain via weakly supervised learning. Zhang \etal \cite{zhang2020inference} propose an online adaptation to target test data based on the weakly supervised learning method of \cite{Chen_2019_CVPR}. Yang \etal\cite{Yang_2018_CVPR} use unpaired 3D annotation to further fine-tune a network on in-the-wild images. Kundu \etal \cite{Kundu_2020_WACV} use a self-supervised learning method to improve the generalization of a pre-trained network on images with occlusion.  

\textbf{Cross-dataset Generalization.} Cross-dataset adaption of 3D pose estimators has recently gained attention. Guan \etal and Zhang \etal ~\cite{Guan_2021_CVPR, zhang2020inference} propose an online adaptation of the pose estimator during the inference stage over test data. Guan \etal ~\cite{Guan_2021_CVPR} use a temporal consistency loss and a 2D projection loss on the streaming test data to adapt the network to the target test dataset. Zhang \etal~\cite{zhang2020inference} use a cycle consistency approach to optimize the network on every single test frame. Although the online-adaptation approach improves cross-dataset generalizability, it also increases the inference time, especially if the networks exploit temporal information. Wang \etal~\cite{Viewpoint_ECCV2020} argues that estimating the camera viewpoint beside the 3D keypoints improves cross-dataset generalization of the 3D pose estimator. However, the camera viewpoint is not the only criterion that differs between datasets. Split-and-Recombine ~\cite{SRnet_ECCV2020} proposes to split the human skeleton into different body parts so that different body parts of a rare pose from the target dataset could have been seen in the source dataset.

\textbf{Data Augmentation} is another way to diminish cross-dataset errors. Previous methods perform data augmentation on images \cite{MoCap-Guided}, 3D mesh models \cite{Deep3DPose,zhang2021bmp,Varol_2017_CVPR}, or 2D-3D pairs \cite{Li_2020_CVPR,Cheng_2019_ICCV,Guan_2021_CVPR}. Most related to our work is augmenting 2D-3D pairs. Li \etal \cite{Li_2020_CVPR} generate synthetic 3D human samples by substituting body parts from a source training set. The evolutionary process of \cite{Li_2020_CVPR} is successful in generating new poses, however, the generation of natural camera viewpoints is overlooked. Instead, it randomly perturbs source camera poses. PoseAug \cite{Gong_2021_CVPR} proposes an end-to-end data augmentation framework that trains along with a pose estimator network. Although it improves the diversity of the training data, there is no guarantee that the generated samples are in the distribution of the target dataset. Moreover, according to the ablation studies of PoseAug, the main improvement comes from generating camera viewpoints instead of generating new poses. This means that PoseAug has limited abilities to effectively improve pose diversities in the training set. 
In contrast, we enforce the generated synthetic data to be in the distribution of the target data. 
Unlike PoseAug, we show that our motion generation network significantly improves cross-dataset results even without augmenting the camera-viewpoints.

\section{Problem Formulation}
Let $\mathbf{X}^{\text{src}}=(X^{\text{src}}_{2D},X^{\text{src}}_{3D})$ be a pair of 2D and 3D poses from the source dataset and $\mathbf{X}^{\text{tar}}=X^{\text{tar}}_{2D}$ a 2D pose from the target dataset. 
The input to our model are sequences of frames with length $n$,
$X^{\text{src}}_{2D}:[x_{2D}]_{t=0}^{n}$, $X^{\text{src}}_{3D}:[x_{3D}]_{t=0}^{n}$, and $X^{\text{tar}}_{2D}:[y_{2D}]_{t=0}^{n}$ where $x_{2D},y_{2D} \in \mathbf{R}^{J \times 3}$. 
AdaptPose consists of a generator function 
\begin{equation}
 {G}(\mathbf{X}^{\text{src}}, \mathbf{z}; \boldsymbol{\theta}_{G})\rightarrow \mathbf{X}^{\text{fake}},   
\end{equation}
with parameters $\boldsymbol{\theta}_{G}$, that maps source samples $\mathbf{X}^{\text{src}}$ and a noise vector $\mathbf{z}\sim p_{z}$ to a fake 2D-3D pair $\mathbf{X}^{\text{fake}}=(X^{\text{fake}}_{2D},X_{3D}^{\text{fake}})$. 
The fake samples $(X_{2D}^{\text{fake}},X_{3D}^{\text{fake}})$ are a sequence of 2D-3D keypoints $X^{\text{fake}}_{2D}:[x^{\text{fake}}_{2D}]_{t=0}^{n}$, $X_{3D}^{\text{fake}}:[x_{3D}^{\text{fake}}]_{t=0}^{n}$. The generator $\mathbf{G}$ generates an adapted dataset $\mathbf{X}^{\text{fake}}={G}(\mathbf{X}^{\text{src}},\mathbf{z})$ of any desired size. In order to adapt the source to the target domain in the absence of 3D target poses we introduce a domain discriminator $D_{D}$ and a 3D discriminator $D_{3D}$. The domain discriminator $D_{D}(\mathbf{x};\boldsymbol{\theta}_{D})$ gives the likelihood $d$ that the 2D input $\mathbf{x}$ is sampled from the target domain $X_{2D}^{\text{tar}}$. The generator tries to generate fake samples $X^{\text{fake}}_{2D}$ as close as possible to target samples $X^{\text{tar}}_{2D}$, while the discriminator tries to distinguish between them. Unlike a standard GAN network \cite{GAN} where generator is conditioned only on a noise vector, our generator is conditioned on both a noise vector and a sample from the source dataset which was shown to be effective in generating synthetic images \cite{Bousmalis_2017_CVPR}. Additionally, the model is conditioned on a 3D discriminator $D_{3D}(\mathbf{x};\mathbf{\theta}_{D})$ that outputs the likelihood $d^{'}$ that the generated 3D, $X_{3D}^{\text{fake}}$, is sampled from the real 3D distribution. Ideally, we would like to condition on the target 3D dataset.
Since 3D data from the target domain is not available we condition it on the source 3D dataset. However, conditioning the 3D discriminator $D_{3D}$ directly on the source 3D poses restrains the motion generator to the source distribution. Instead, we condition the 3D discriminator $D_{3D}$ on a perturbed version of data ${X}^{\text{psrc}}_{3D}=\mathbf{y}+{X}^{\text{src}}_{3D}$ where $\mathbf{y}\sim p_{y}$ is a small noise vector. The noise vector $\mathbf{y}$ is selected such that ${X}^{\text{psrc}}_{3D}$ is a valid pose from the source distribution. The goal of AdaptPose is to optimize the following objective function
\begin{equation}
    \mathcal{L}=\underset{\boldsymbol{\theta}_{G}}{{\text{\normalsize min  }}}
    \underset{(\boldsymbol{\theta}_{D_{D}}, \boldsymbol{\theta}_{D_{3D}})}{{\text{\normalsize max  }}} \mathcal\alpha{L}(G,D_{D})+\beta\mathcal{L}(G,D_{3D})
    ,
\end{equation}
where $\alpha$ and $\beta$ are the weights of the losses.

\section{Human Motion Generator}
\label{sec:HMG}

We name the generator of our GAN network \textit{Human Motion Generator} (HMG). The HMG consists of two main components. 1) A \textit{bone generator} that rotates the bone vectors and changes the bone length ratios. The bone generation operation produces new 3D keypoints $X^{b}_{3D}$. 2) A \textit{camera generator} that generates a new camera viewpoint $\{\textbf{R},\textbf{T}\}$, where $\textbf{R}\in \mathbb{R}^{3\times 3}$ is a rotation matrix and $\textbf{T}$ is a translation vector. $X^{b}_{3D}$ is transformed to the generated camera viewpoint by
 \begin{equation}
{X}^{\text{fake}}_{3D}=\textbf{R}{X}^{b}_{3D}+\textbf{T},
  \label{eq:important}
\end{equation}
with the corresponding 2D keypoints
 \begin{equation}
{X}^{\text{fake}}_{2D}=\Pi (X^{\text{fake}}_{3D}),
  \label{eq:important}
\end{equation}
where $\Pi$ is the perspective projection that uses the intrinsic parameters from the source dataset.

\subsection{Bone Generation}
\label{subsec:BG}
\begin{figure}[t]
\begin{center}
% \fbox{\rule{0pt}{2in} \rule{.4\linewidth}{0pt}}
\includegraphics[scale=0.5]{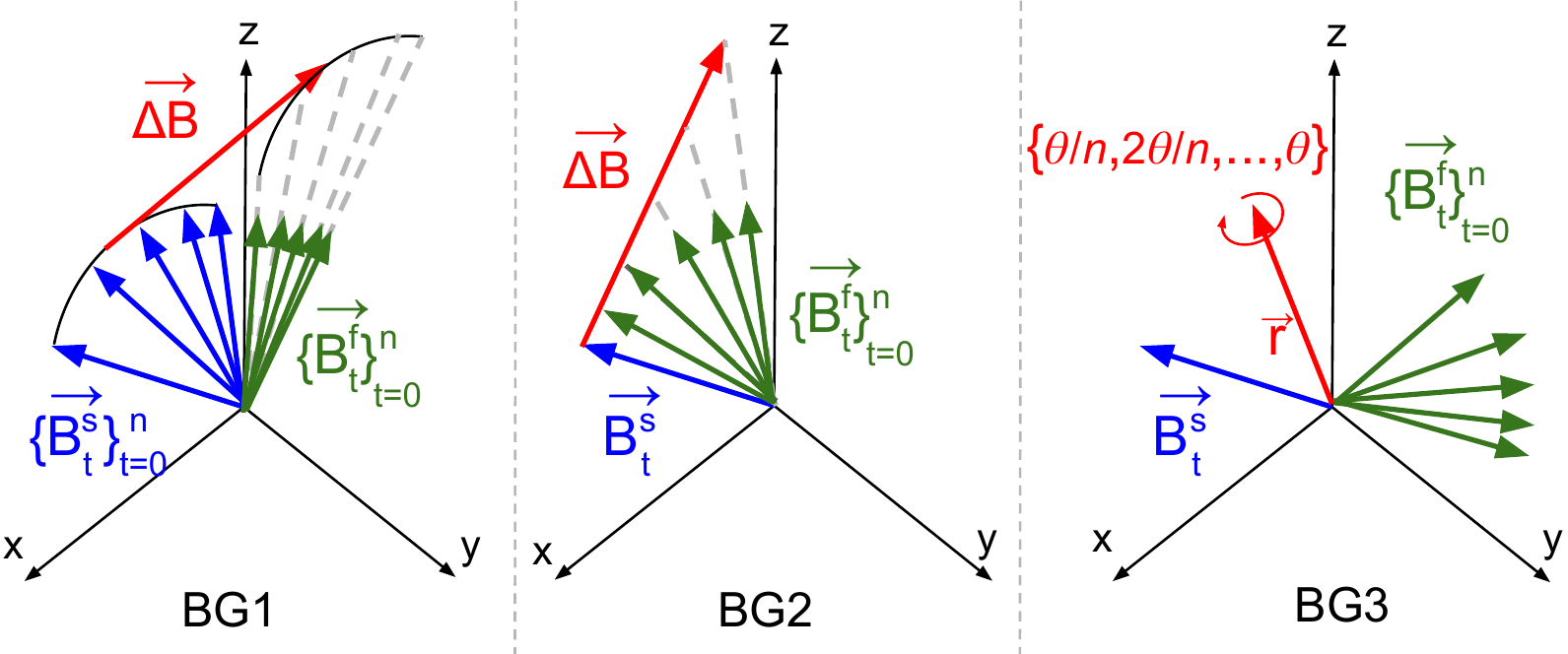}
\end{center}
  \caption{Bone generation methods. Blue vectors indicate bone vectors before rotation and green vectors are bone vectors after rotation. $\Delta \vec{B}$ is rotating bone direction produced by the network. $\vec{r}$ and $\theta$ are the axis and angle of the rotation, respectively.}
\label{fig:BoneGeneration}
\end{figure}

In this section, we analyze different methods of bone vector generation in the temporal domain. The main challenge is to keep the bone changes plausible for every single frame and temporally consistent in the time domain. We propose and analyze the three different methods BG1, BG2, and BG3 shown in Figure~\ref{fig:BoneGeneration}. 

\textbf{BG1.} The bone generation network accepts a sequence of 3D keypoints from the source dataset. The sequence of 3D keypoints is transformed into a bone vector representation $[\vec{B}_{t}^{\text{src}}]_{t=t0}^{t0+n}$ where $\vec{B}_{t}^{\text{src}} \in \mathbb{R}^{(J-1) \times 3}$ and $J$ is the number of keypoints. BG1 generates a displacement vector $\Delta \vec{B} \in \mathbb{R}^{(J-1) \times 3}$ and a bone ratio $\lambda \in \mathbb{R}^{(J-1) \times 1}$. The new bone vector is $[\vec{B}_{t}^{\text{fake}}]_{t=t0}^{t0+n}$ where
\begin{equation}
\vec{B}^{\text{fake}}_{t}=\frac{\vec{B}^{\text{src}}_{t}+\Delta \vec{B}}{\|\vec{B}^{\text{src}}_{t}+\Delta \vec{B}\|}\|{\vec{B}^{\text{src}}_{t}}\|(1+\lambda).
  \label{eq:BG1}
\end{equation}
$\Delta \vec{B}$ may change the bone length instead of rotating to a new configuration as shown in 
%BG1 of 
Figure \ref{fig:BoneGeneration}. 
%Therefore 
To avoid this, we divide the generated bones by $\| \vec{B}^{\text{src}}_{t}+\Delta \vec{B}\|$ in Eq.~\ref{eq:BG1}.
 
\textbf{BG2.} The bone generation network accepts a single sample of 3D keypoints from the source dataset and converts it to a bone representation $\vec{B}_{t0}^{\text{src}}$. BG2 generates $\Delta \vec{B}$ and $\lambda$. The new bone vector is $[\vec{B}_{t}^{\text{fake}}]_{t=t0}^{t0+n}$ where
\begin{equation}
\vec{B}^{\text{fake}}_{t+j}=\frac{\vec{B}^{\text{src}}_{t0}+\textit{j}\Delta \vec{B}/n}{\|{\vec{B}^{\text{src}}_{t0}+\textit{j}\Delta \vec{B}/n}\|}\|{\vec{B}_{t0}^{\text{src}}}\|(1+\lambda).
  \label{eq:BG2}
\end{equation}
\textbf{BG3.} The bone generation network generates the vector $\vec{r}\in \mathbb{R}^{(J-1) \times 3}$ and the angle $\theta \in \mathbb{R}^{(J-1) \times 1}$. A sequence of rotation matrices $[R_{t}]_{t=0}^{n}$ is calculated by
\begin{equation}
R_{t+j}=\mathcal{H}(\frac{\vec{r}}{\norm{\vec{r}}}\frac{\textit{j}\theta}{n}),
  \label{eq:BG3}
\end{equation}
where $\mathcal{H}$ transforms axis-angle rotation of $(\theta, \vec{r})$ to rotation matrix representation via quaternions $q=q_r+q_x\textbf{i}+q_y\textbf{j}+q_z\textbf{k}$ by
\begin{equation}
q=\cos(\frac{\theta}{2})+\frac{\vec{r}}{\norm{\vec{r}}}\sin(\frac{\theta}{2}),
  \label{eq:quat_to_rot}
\end{equation}
\begin{equation}
R=v\otimes v+q_r^2\textbf{I}+2q_r[v]_{\times}+[v]_{\times}^{2},
  \label{eq:important}
\end{equation}
where $\otimes$ is the outer product, $\textbf{I}$ is the identity matrix, and 
\begin{equation}
    [v]_{\times}=\begin{bmatrix}
    0&-v_{3}&v_{2}\\
    v_{3}&0&-v_{1}\\
    -v_{2}&v_{1}&0
    \end{bmatrix}.
\end{equation}
\subsection{Camera Generation}
\label{subsec:CG}
In this section, we introduce two different methods of camera generation: \textit{1) Deterministic,} which generates a single camera rotation matrix and translation and \textit{2) probabilistic.} The network learns a distribution of rotation matrices. A random rotation matrix is sampled from the learned distribution. 
Additionally, we explore three different rotation representations: axis-angle, Euler-angles, and quaternions. In the following, we will discuss each of the procedures for each of the rotation representations.

\textbf{Deterministic Axis-angle.} The network generates an axis $\vec{r}$ and a translation $T$ where the angle of rotation is $\|\vec{r}\|$. The rotation matrix $R\in \mathbb{R}^{3\times3}$ is produced by $R=\mathcal{H}(\vec{r})$ where 
$\mathcal{H}$ is explained in the equation \ref{eq:quat_to_rot}.

\textbf{Probabilistic Axis-angle.} The network learns three separate normal distributions $\mathcal{N}_{1} (\mu_1,\sigma_1)$, $\mathcal{N}_{2} (\mu_2,\sigma_2)$ , and $\mathcal{N}_{3} (\mu_3,\sigma_3)$, an angle $\theta$, and a translation $T$. The axis $r=\{r_1,r_2,r_3\}$ is sampled from the learned normal distributions and converted to a rotation matrix by
\begin{equation}
   R=\mathcal{H}(\frac{\vec{r}}{\norm{\vec{r}}} \theta).
\end{equation}
\textbf{Probabilistic Euler-angles.} The network learns three Gaussian distributions $\mathcal{N}_{1}$, $\mathcal{N}_{2}$, and $\mathcal{N}_{3}$ to sample the Euler-angles $(\alpha, \beta, \gamma)$
from the specified distributions. The rotation matrix is obtained as follows:
\begin{equation}
    R=R_{z}(\alpha)R_{y}(\beta)R_{x}(\lambda),
\end{equation}
where $R_{z}(\alpha),R_{y}(\beta),$ and $R_{x}(\lambda)$ are rotations of $(\alpha, \beta,$ and  $\gamma)$ degrees around $z,y, x$ axis, respectively.

\textbf{Probabilistic Quaternion.} A quaternion represents a rotation around axis $\vec{u}=(u_x,u_y,u_z)$ with angle $\theta$ as 
\begin{equation}
    q=\cos(\frac{\theta}{2})+\vec{u}\sin(\frac{\theta}{2}).
\end{equation}
Therefore, $q$ can be represented by four elements. Our network learns four distributions $\mathcal{N}_{1,\dots,4}$ and randomly samples elements of $q$ from the distributions. The quaternion $q$ is then converted to a rotation matrix representation as explained in section \ref{subsec:BG}.
\subsection{Domain and 3D Discriminators}
\label{subsec:DD}
We adopt the kinematic chain space (KCS)\cite{Wandt_2019_CVPR,Wandt_2018_ECCV_Workshops} in 2D space to generate a matrix of joint angles and limb lengths in the image plane. The domain discriminator has two branches that accept 2D keypoints and the KCS matrix, respectively. The diagonal of the KCS matrix contains the limb lengths in the image space. Other components of the KCS matrix represent angular relationships of the 2D pose. It is important to mention that we do not normalize input 2D keypoints relative to the root joint as it causes perspective ambiguities \cite{Yu_2021_CVPR}. Therefore, $diag(KCS)$ is a function of position and body scale. On the contrary $KCS-diag(KCS)$ is a function of the camera viewpoint and scale of the person. Thus, the KCS matrix disentangles different parameters that the motion generator requires to learn. For the 3D discriminator, in order not to condition the 3D discriminator on the source domain, we first apply a random perturbation of $\beta$ degrees to the input bone vectors $\beta < 10^{\circ}$  and then feed the perturbed 3D to a part-wise KCS branch \cite{Gong_2021_CVPR} and the original 3D to a KCS branch. Further details about the 3D discriminator are provided in the supplementary material.
\subsection{Selection}
In order to stabilize the training of the lifting network 
%with very hard synthesized samples 
we introduce a selection step by evaluating samples via the lifting network $N$. In this step, the lifting network receives $(X_{2D}^{\text{src}},X_{3D}^{\text{src}})$ and $(X_{2D}^{\text{fake}},X_{3D}^{\text{fake}})$ which are source and generated samples, respectively. We exclude samples that are either too simple or too hard using the following rule
\begin{equation}
\text{selection=}
\begin{cases}
    \text{yes}& \text{if  }(\frac{\mathcal{L}(N(X_{2D}^{\text{fake}}))}{\mathcal{L}(N(X_{2D}^{\text{src}}))}-a)^{2}<b^{2}\\
    \text{no}&\text{otherwise}
\end{cases}
,
\end{equation}
where $\mathcal{L}$ is an $L_{2}$ loss.
\section{Training}

\textbf{Motion Generator.}  Our adversarial framework is trained using three losses for the motion generator and for the discriminators which are defined as
\small
\begin{align}
\centering
\mathcal{L}_{\mathcal{D}_{3D}} &= \frac{1}{2} \mathbb{E}[(\mathcal{D}(X_{3D}^{\text{src}})-1)^{2}] +\frac{1}{2} \mathbb{E}[\mathcal{D}(X^{\text{fake}}_{3D})^{2}], \\
\mathcal{L}_{\mathcal{D}_{D}} &=\frac{1}{2} \mathbb{E}[(\mathcal{D}(X_{2D}^{\text{src}})-1)^{2}] +\frac{1}{2} \mathbb{E}[\mathcal{D}(X^{\text{fake}}_{2D})^{2}], \\
\mathcal{L}_{{G}_{adv}} &=\frac{1}{2} \mathbb{E}[(\mathcal{D}(X^{\text{fake}}_{2D})-1)^{2}], 
\label{equ:wasserstein_loss}
\end{align}
\normalsize
where $(X^{\text{src}}_{3D}, {X}^{\text{fake}}_{3D})$ are 3D samples from the source dataset and synthetic generated samples, respectively. $(X^{\text{tar}}_{2D}, X^{\text{fake}}_{2D})$ are 2D keypoints from the target dataset and the generated synthetic data, respectively. The generator also receives a feedback loss from the lifting network. The feedback loss has two components: 1) reprojection loss of the estimated 3D keypoints of the target domain 2) fixed hard ratio feedback loss adapted from \cite{Gong_2021_CVPR}. The lifting network $N$ accepts $X_{2D}^{\text{tar}}$ from the target dataset and predicts $X_{3D}^{\text{tar}}$. 
We define the reprojection loss as
\begin{equation}
    \mathcal{L}_{proj}=\norm{ \frac{X_{proj}^{\text{fake}}}{{\|X_{proj}^{\text{fake}}\|}}-\frac{X_{2D}^{\text{fake}}}{{\|X_{2D}^{\text{fake}}}\|}}_{1},
\end{equation}
where $\norm{}_{1}$ is the $L_1$ norm and
\begin{equation}
 X_{proj}^{\text{fake}} =
 \begin{bmatrix}  
  1 & 0 & 0 \\
  0 & 1 & 0 
  \end{bmatrix} {N}(X_{2D}^{\text{tar}}).
\end{equation}The fixed hard ratio loss provides feedback depending on the difficulty of generated sample relative to the source samples as follows:
\begin{equation}
    f=(\frac{\mathcal{L}(N({X_{2D}^{\text{fake}}}))}{\mathcal{L}(N({X_{2D}^{\text{src}}}))}-c)^{2},
\end{equation}

\begin{equation}
  \mathcal{L}_{hr} =
    \begin{cases}
      f & \text{if  } f<d^{2}\\
      0 & \text{otherwise}
    \end{cases},       
\end{equation}
where $\mathcal{L}$ is $L_{2}$ loss. The summation of the above mentioned losses is our generator loss
\begin{equation}
    \mathcal{L}_{G}=\mathcal{L}_{advG}+\mathcal{L}_{proj}+\mathcal{L}_{hr}.
\end{equation}

\textbf{Lifting Network.} The lifting network $N$ is trained using $(X_{2D}^{\text{src}},X_{3D}^{\text{src}})$ and $(X_{2D}^{\text{fake}},X_{3D}^{\text{fake}})$
which gives the lifting loss
\begin{equation}
\small
    \mathcal{L}_{N}=\norm{X_{3D}^{\text{src}}-N(X_{2D}^{\text{src}})}_{2}+\norm{X_{3D}^{\text{fake}}-N(X_{2D}^{\text{fake}})}_{2}.
\end{equation}

\section{Experiments} 
We perform extensive experiments to evaluate the performance of AdaptPose for cross-dataset generalization. We further conduct ablation studies on the different elements of our network. In the following, we discuss different datasets and subsequently baselines and metrics.

\begin{itemize}
    \item \textbf{Human3.6M (H3.6M)} contains 3D and 2D data from seven subjects captured in 50 fps. 
    We use the training set of H3.6M (S1, S5, S6, S7, S8) as our source dataset for cross-dataset evaluations. 
    While performing experiments on the H3.6M dataset itself we will use S1 as the source dataset and S5, S6, S7, and S8 as the target. 
    % Since most of the datasets are captured in a frame rate lower than 50fps, we also perform random downsampling in our data loader which we will discuss further in the baseline section. All the actions are recorded in a controlled indoor environment.   
    
    \item \textbf{MPI-INF-3DHP (3DHP)} contains 3D and 2D data from 8 subjects and covers 8 different activities. 
    % 3DHP includes fast motions including Boxing, tennis, golf, soccer. 
    We will use the 2D data from the training set of 3DHP \cite{mono-3dhp2017} as our target dataset when evaluating 3DHP. 
    % The test set of 3DHP includes data from outdoor activities as well. 
    The test set of 3DHP includes more than 24K frames. However, some of the previous work use a subset of test data which includes 2,929 frames for evaluation \cite{Gong_2021_CVPR,kolotouros2019spin}. The 2,929 version has temporal inconsistency which is fine for the single-frame networks. We use the official test set of 3DHP and compare our results against the previous work's results on the official test set of 3DHP for a fair comparison.
    
    \item\textbf{3DPW} contains 3D and 2D data captured in an outdoor environment. The camera is moving in some of the trials. 3DPW \cite{vonMarcard2018} is captured in 25fps and has more variability than 3DHP and H3.6M  in terms of camera poses. We use the training set of 3DPW as our target dataset when experimenting on this dataset. 
    \item \textbf{Ski-Pose PTZ-Camera (Ski)} includes 3D and 2D labels from 5 professional ski athletes in a ski resort. The dataset is captured in 30 fps and frames are cropped in $256\times256$. The cameras are moving and there is a major domain gap between Ski and previous datasets in terms of the camera pose/position. 
\end{itemize}    

\textbf{Evaluation Metrics.}
We use mean per joint position error (MPJPE) and Procrustes aligned MPJPE (P-MPJPE) as our main evaluation metrics. P-MPJPE measures MPJPE after performing Procrustes alignment of the predicted pose and the target pose. We also report the percentage of correct keypoint (PCK) with a threshold of 150 mm and area under the curve (AUC) for evaluation on 3DHP following previous arts.

\textbf{Baseline.} We use VideoPose3D\cite{pavllo:videopose3D:2019} (VPose3D) as the baseline pose estimator model. VPose3D is a lifting network that regresses 3D keypoints from input 2D keypoints. We use 27 frames as the input in our experiments. As preprocessing for H3.6M, 3DHP, and 3DPW datasets we normalize image coordinate such that $[0,w]$ is mapped to $[-1,1]$. 
Note that the 3DPW dataset has some portrait frames with a height greater than width. In these cases, we pad the width so that height is equal to width to avoid the 2D keypoint coordinates being larger than the image frame after normalization. 
Our experiments show that this preprocessing has lower cross-dataset error compared with root centering and Frobenius normalization of 2D keypoints.  While performing experiments on the Ski dataset we use root centering and Frobenius normalization of 2D keypoints since the image frames are already cropped to $256\times256$ with the person in the center of the image. Since there is an fps difference and also motion speed difference between our source dataset and target datasets, we also perform random downsampling in our data loader for training the baseline network. Specifically, our data loader samples $\{x_{r(t-n)},...,x_{r(t+n)}\}$ from the source dataset and $r$ is a random number sampled from a uniform distribution of $[2,5]$ . Table \ref{tab:Ablation_model} shows that the baseline model has a cross-dataset MPJPE of 96.4 mm using 3DHP as the target dataset.

\subsection{Quantitative Evaluation}
\textbf{H3.6M.} We compare our results with previous semi-supervised learning methods that only use 3D labels from S1 and 2D annotations from the remaining subjects for training \cite{pavllo:videopose3D:2019} as well as data augmentation methods. Our results improve upon the previous state-of-the-art by 16\%. We use ground truth 2D keypoints and therefore compare with previous work with the same setting. Since the camera pose does not change much between subjects, we hypothesize that the comparison in the current setting compares our bone generation method against previous work. 
% Table \ref{tab:h36m} shows that PoseAug \cite{Gong_2021_CVPR} has larger error than \cite{Li_2020_CVPR} in this setting even-though PoseAug is superior in cross-dataset evaluations.

\textbf{3DHP.} Table \ref{tab:3DHP} gives MPJPE, AUC, and PCK on test set of 3DHP. We report the results of PoseAug's released pre-trained model on the complete test set of 3DHP. Our results have a 14\% margin in terms of MPJPE compared with previous methods that report cross-dataset evaluation results \cite{Li_2020_CVPR,Gong_2021_CVPR, Guan_2021_CVPR, Viewpoint_ECCV2020, SRnet_ECCV2020}. This includes the comparison to \cite{zhang2020inference} that uses information from the target test data to perform test-time optimization.
% Our results are even better than models fine-tuned with 3D GT of 3DHP which highlights the adaptation approach for generalization of pre-trained models. 

\textbf{3DPW.} Table \ref{tab:3DPW} provides MPJPE and PA-MPJPE on the test set of 3DPW. Our method outperforms previous methods by 12 mm in PA-MPJPE. This includes previous methods that particularly were designed for cross-dataset generalization \cite{Guan_2021_CVPR, Gong_2021_CVPR, Sim2real} and those that use temporal information \cite{Kocabas_2019_CVPR,Guan_2021_CVPR}. In comparison with test-time optimization methods \cite{Guan_2021_CVPR, zhang2020inference}, ours also has an advantage of fast inference.

\textbf{SKI.} Table \ref{tab:SKI} gives the cross-dataset results on the Ski dataset. Skiing is fast and sequences of the Ski dataset are as short as 5s. This provides little training data for temporal models and, therefore, we use a single-frame input model. We report the performance of VPose3D with single-frame input in a cross-dataset scenario to compare as a baseline model. Moreover, our results compared with Rhodin \etal \cite{Rhodin_2018_CVPR} and CanonPose \cite{Wandt_2021_CVPR} that use multi-view data from the training set of Ski show 28mm improvement in MPJPE and 2mm in PA-MPJPE. 
%-------------------------------------------------------------------------
\begin{table}
\centering

\caption{Cross-scenario learning on H3.6M. Source: S1. Target: S5, S6, S7, S8}
\label{tab:h36m}
\small
\begin{tabular}{ p{2.8cm}|c|ccc}
\hline
Method&3D&PA-MPJPE&MPJPE\\
\hline
Martinez \textit{et al.}\cite{Martinez_2017_ICCV}&Full&--&45.5\\
Pavllo\cite{pavllo:videopose3D:2019}&Full&27.2&37.2\\
Lui \textit{et al} \cite{Liu_2020_CVPR}&Full&--&34.7\\
Wang \cite{Wang_ECCV2020_motion_guided}&Full&-&25.6\\
\hline
PoseAug\cite{Gong_2021_CVPR}&S1&--&56.7\\
Pavllo\cite{pavllo:videopose3D:2019}&S1&--&51.7\\
Li \textit{et al.}\cite{Li_2020_CVPR}&S1&--&50.5\\\hline
Ours&S1&\textbf{34.0}&\textbf{42.5}\\
\hline
\end{tabular}
\end{table}

\begin{table}
\small
\centering

\caption{Cross-dataset evaluation on 3DHP dataset. Source: H3.6M-target:3DHP}
\label{tab:3DHP}
\begin{tabular}{ p{2.4cm}|c|ccc}
\hline
Method&CD&PCK&AUC&MPJPE\\
\hline
Mehta \textit{et al.}\cite{mono-3dhp2017}&&76.5&40.8&117.6\\
VNet\cite{VNect_SIGGRAPH2017}&&76.6&40.4&124.7\\
MultiPerson\cite{Multi-person}&&75.2&37.8&122.2\\
OriNet \cite{Luo2018OriNetAF}&&81.8&45.2&89.4\\
\hline
BOA\cite{Guan_2021_CVPR}&\checkmark&\textbf{90.3}&-&117.6\\
Wang \textit{et al.}\cite{Viewpoint_ECCV2020}&\checkmark&76.1&-&109.5\\
SRNET\cite{SRnet_ECCV2020}&\checkmark&77.6&43.8&-\\
Li \textit{et al.}\cite{Li_2020_CVPR}&\checkmark&81.2&46.1&99.7\\
% Guang \textit{et al.}\cite{Guan_2021_CVPR}&\checkmark&90.3&-&117.6\\
PoseAug\cite{Gong_2021_CVPR}&\checkmark&82.9&46.5&92.6\\
Zhang \textit{et al.}\cite{zhang2020inference}&\checkmark&83.6&48.2&92.2\\
\hline
Ours&\checkmark&88.4&\textbf{54.2}&\textbf{77.2}\\
\hline
\end{tabular}
\end{table}

\begin{table}
\small
\centering
\caption{Cross-dataset evaluation on 3DPW dataset. Source: H3.6M-target:3DPW}
\label{tab:3DPW}
\begin{tabular}{ p{2.4cm}|c|cc}
\hline
Method&CD&PA-MPJPE&MPJPE\\
\hline
EFT\cite{joo2021exemplar}&&55.7&--\\
Vibe\cite{Kocabas_2020_CVPR}&&51.9&82.9\\
Lin \textit{et al.}\cite{Lin_2021_ICCV}&&45.6&74.7\\
\hline
Sim2real\cite{Sim2real} &\checkmark&74.7&--\\
Zhang \textit{et al.}\cite{zhang2020inference}&\checkmark&70.8&--\\
Wang \textit{et al.}\cite{Viewpoint_ECCV2020}&\checkmark&68.3&109.5\\
SPIN \cite{kolotouros2019spin}&\checkmark &59.2&96.9\\
PoseAug\cite{Gong_2021_CVPR}&\checkmark&58.5&94.1\\
VIBE \cite{Kocabas_2020_CVPR}&\checkmark&56.5&93.5\\
BOA \cite{Guan_2021_CVPR}&\checkmark&49.5&\textbf{77.2}\\\hline
Ours&\checkmark&\textbf{46.5}&81.2\\
\hline
\end{tabular}
\end{table}

\begin{table}
\small
\centering
\caption{Cross-dataset evaluation on Ski dataset. Source: H3.6M-target:Ski}
\label{tab:SKI}
\begin{tabular}{ p{2.4cm}|c|cc}
\hline
Method&CD&PA-MPJPE&MPJPE\\
\hline
Rhodin \etal\cite{Rhodin_2018_CVPR}&&85&--\\
CanonPose\cite{Wandt_2021_CVPR}&&89.6&128.1\\
\hline
Pavllo \etal \cite{pavllo:videopose3D:2019}&\checkmark&88.1&106.0\\
PoseAug \cite{Gong_2021_CVPR}&\checkmark&83.5&105.4\\
Ours&\checkmark&\textbf{83.0}&\textbf{99.4}\\
\hline
\end{tabular}
\end{table}
\subsection{Qualitative Evaluation}
\begin{figure}[t]
\setlength{\belowcaptionskip}{-10pt}
\begin{center}
% \fbox{\rule{0pt}{2in} \rule{.4\linewidth}{0pt}}
\includegraphics[scale=0.4]{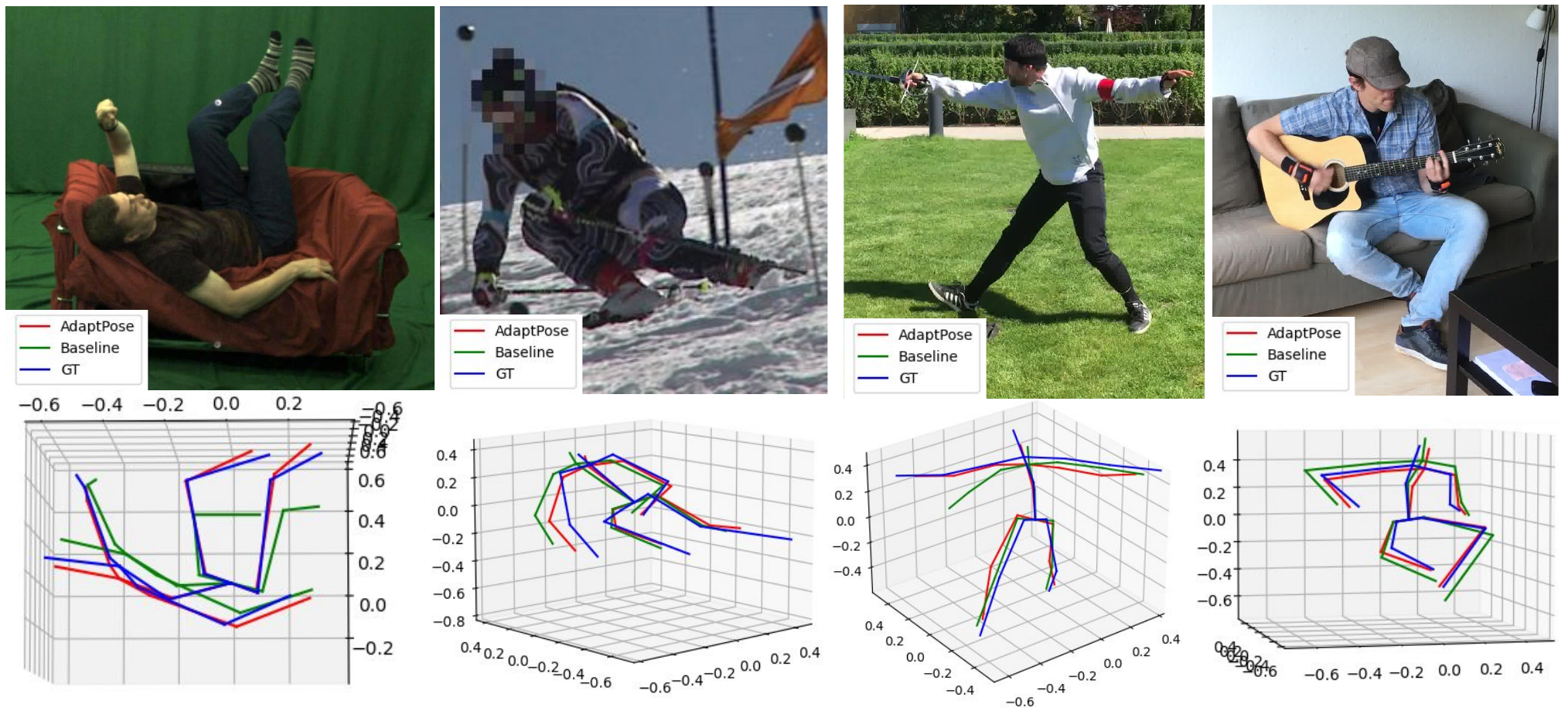}
\end{center}
  \caption{ 3D human pose predictions (red) vs. ground truth (blue) for samples of Ski and 3DPW.}

\label{fig:Qualitative_pred}
\end{figure}

\begin{figure}[t]
\setlength{\belowcaptionskip}{-10pt}
\begin{center}
% \fbox{\rule{0pt}{2in} \rule{.4\linewidth}{0pt}}
\includegraphics[scale=0.27]{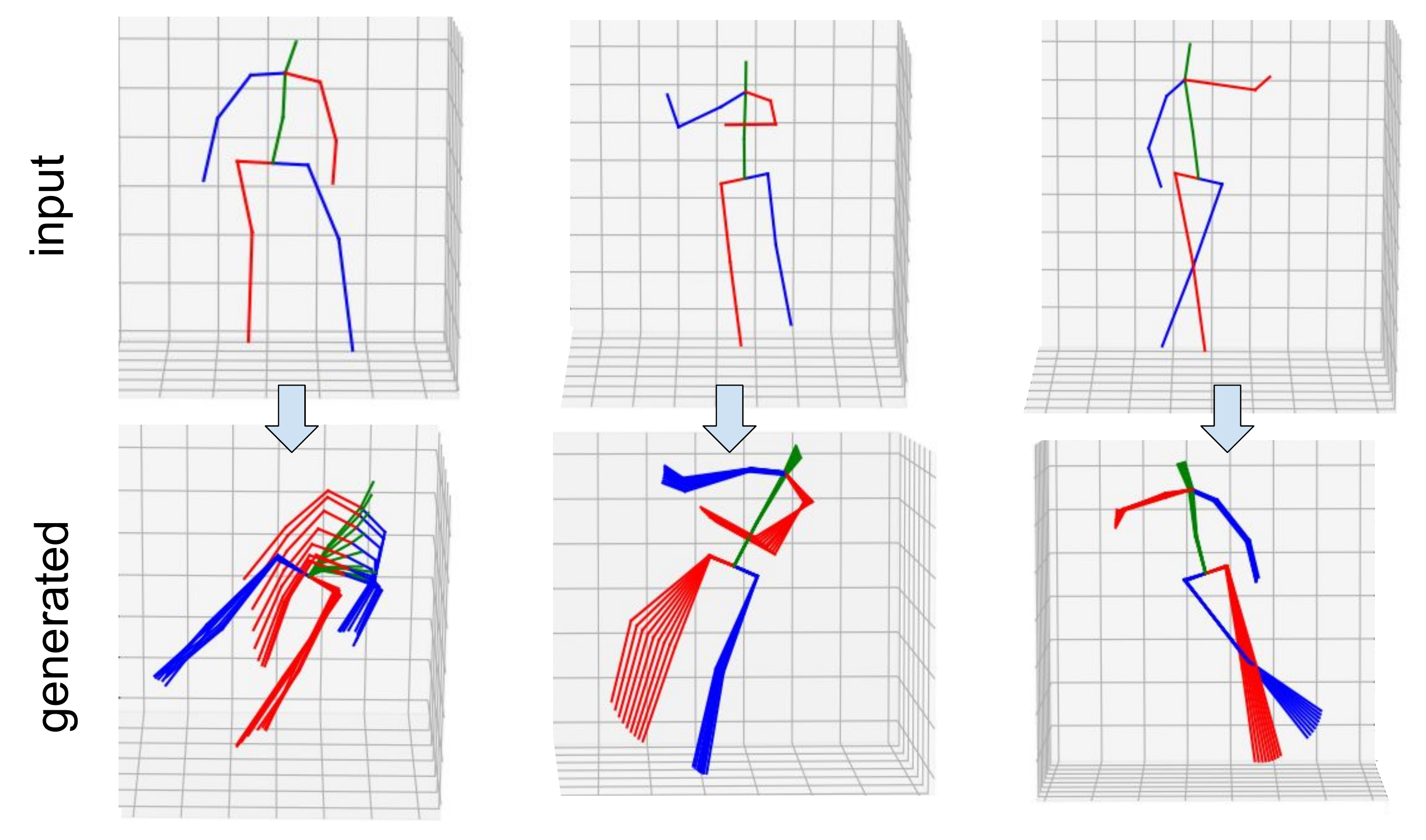}
\end{center}
  \caption{Samples of generated motions and the corresponding input 3D keypoints. Motions are smooth and realistic. }

\label{fig:motion}
\end{figure}

Figure \ref{fig:Qualitative_pred} shows qualitative evaluation on Ski, 3DHP, and 3DPW datasets. The predictions of the baseline and AdaptPose are depicted vs. the ground truth. We observe that AdaptPose successfully enhances the baseline predictions. Figure \ref{fig:motion} provides some examples of the generated motion and the input 3D keypoints. Generated motions are smooth and realistic. We provide further qualitative examples in the supplementary material.

\subsection{Ablation Studies}

\textbf{Ablation on Components of AdaptPose.} We ablate components of our framework including bone generation, camera generation, domain discriminator, and selection. Table \ref{tab:Ablation_model} provides the performance improvements by adding any of the components starting from the baseline. All of the components have a major contribution to the results. Comparing bone generation and camera generation, the latter has larger effects on the performance. However, in contrast to PoseAug\cite{Gong_2021_CVPR}, our bone generation method is significantly contributing to the results (10 mm vs 1 mm). A3 shows that a combination of bone and camera generation is as good as camera generation alone. Therefore, A4 excludes bone generation from the pipeline that causes a 9 mm performance drop in MPJPE. A3 and A5 give the role of domain adaptation that is 10 mm improvements. 
{\renewcommand{\arraystretch}{1}
\begin{table}[h!]\setlength\tabcolsep{4pt}
\small
\centering
\caption{Ablation study on supervision elements of the proposed model. Source: H3.6M-target:3DHP}
\label{tab:Ablation_model}
\begin{tabular}{ c|cccc|cc}
\hline
Index& BG &Cam & DD  & Select& PMPJPE & MPJPE \\
\hline
\textbf{Baseline}& & &  &&66.5&96.4 \\
\textbf{A1}&\checkmark & & &&61.7 & 90.1\\
\textbf{A2}& & \checkmark& &&62.0 &88.2 \\
\textbf{A3}& \checkmark&\checkmark&  && 61.8&88.1\\
\textbf{A4}& &\checkmark&\checkmark  &\checkmark&59.3 &86.5\\
\textbf{A5} &\checkmark&\checkmark &\checkmark & &54.0 &78.6 \\
\textbf{AdaptPose} &\checkmark &\checkmark & \checkmark & \checkmark&53.6 &77.2\\
\hline
\end{tabular}
\end{table}}

\noindent\textbf{Ablation on bone generation methods.} In this section we compare the performance of three different bone generation methods that were explained in Section \ref{subsec:BG}. Table \ref{tab:Ablation_bone} gives performance of BG1, BG2, and BG3 while performing cross-dataset evaluation on 3DHP. We observe that using an axis-angle representation for rotating bone vectors is superior to generating bone directions. We hypothesize that learning $\Delta \vec{B}$ is a harder task since there are infinitely many $\Delta \vec{B}$ that can generate $[\vec{B}^{'}]_{t=0}^{N}$ from $\vec{B}_{t}$. On the contrary, there are only two axis-angles that map $\vec{B}_{t}$ to $[\vec{B}^{'}]_{t=0}^{N}$.  

{\renewcommand{\arraystretch}{1}
\begin{table}[h!]
\small
\centering
\caption{Ablation study on bone generation strategies}
\label{tab:Ablation_bone}
\begin{tabular}{ p{2cm}|cc}
\hline
Method& PMPJPE & MPJPE \\
\hline
BG1& 59.3& 85.1 \\
BG2&  56.2&80.0\\
BG3& 53.6& 77.2 \\
\hline
\end{tabular}
\end{table}}
\noindent\textbf{Ablation on camera generation methods} In this section we perform analysis on three different camera generation methods that were introduced in Section \ref{subsec:CG}.  In terms of rotation representation, axis-angle outperforms quaternions and Euler-angles. Euler-angles are sensitive to the order of rotations and can lead to degenerate solutions. Comparing probabilistic and deterministic methods, the former obtains 5 mm more accurate results.

{\renewcommand{\arraystretch}{1}
\begin{table}[h!]
\small
\centering
\noindent\caption{Ablation study on camera generation strategies}
\label{tab:Ablation_Cam}
\begin{tabular}{ p{2cm}|c|cc}
\hline
Method& Representation& PMPJPE & MPJPE \\
\hline
Deterministic&Axis-Angle & 58.0& 82.8 \\
Probabilistic&Axis-Angle& 53.6&  77.2\\
Probabilistic&Quaternion& 58.7 & 83.5 \\
Probabilistic&Euler-Angle& 60.9& 85.3 \\
\hline
\end{tabular}
\end{table}}

\noindent\textbf{Ablation on temporal information.} Table \ref{tab:Ablation_temporal} shows the performance of the network while excluding temporal information from the input and generating single 2D-3D pairs. Our cross-dataset MPJPE is 86.4 mm which still improves over previous methods (86.4 mm vs. 92.2). Therefore, although using temporal information is highly contributing to our framework, our network still excels in non-temporal settings.  

{\renewcommand{\arraystretch}{1}
\begin{table}[t]
\small
\centering
\caption{Ablation study on temporal information}
\label{tab:Ablation_temporal}
\begin{tabular}{ p{2cm}|ccc}
\hline
Input& PCK & AUC & MPJPE \\
\hline
1 frame& 84.6 &50.3& 86.4 \\
27 frames& 88.4 &54.2&77.2\\
% 81frames& &  \\
\hline
\end{tabular}
\end{table}}

\subsection{Are we really adapting to new datasets?}
\begin{figure}[t]
\setlength{\belowcaptionskip}{-10pt}
\begin{center}

\includegraphics[scale=0.32]{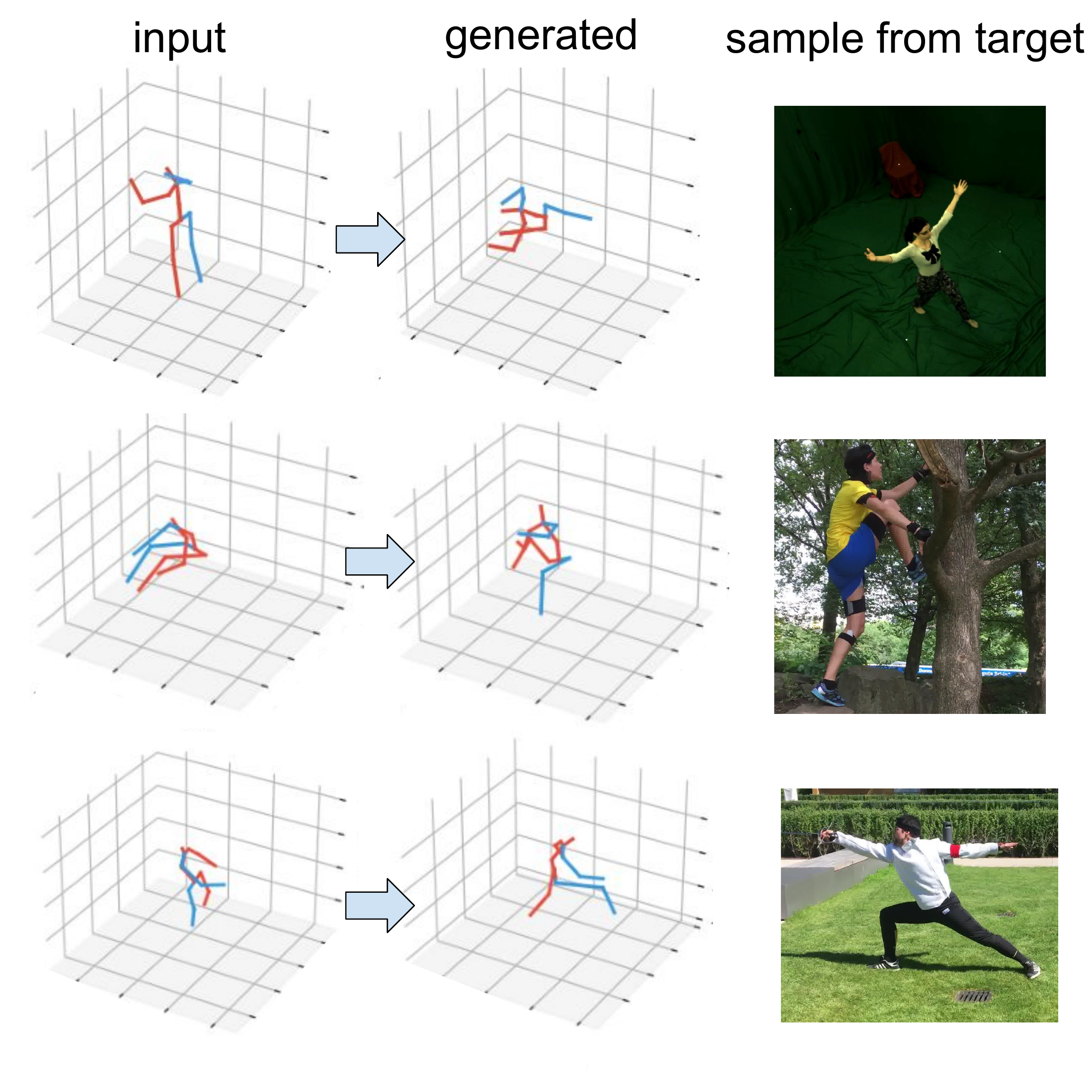}
\end{center}
  \caption{Sample of input images from the source dataset and the generated 3D keypoints. For visualization purposes, we only plot the middle frame from the sequence of generated frames. We manually select the images on the right from the target that matched the generated.}
\label{fig:adapt}
\end{figure}

To evaluate our claim that we are adapting poses and camera views to the target dataset we visualize some samples of generated motions for 3DHP and 3DPW datasets in Figure \ref{fig:adapt}. The ceiling viewpoint in the first row is from 3DHP that is out of the distribution of our source dataset. While the 2D input is from a chest view camera the generated sample is from a ceiling view, similar to the target samples. We observe that our approach generates qualitatively similar camera poses. The second and third rows also provide examples of new poses that are out of the distribution of source poses and similar to samples in the target dataset. We provide further qualitative examples in the supplementary material. Table \ref{tab:Ablation_model} also provides numbers regarding the importance of domain discriminators in our framework (A5 vs A3). It is important to mention that we substitute the domain discriminator with a 2D discriminator from the source dataset when excluding the domain discriminator in Table \ref{tab:Ablation_model}. Thus, the performance drop while excluding the domain discriminator is essentially attributed to the lack of adaptation to the target space and not because of excluding the 2D discriminator. The supplementary material provides further experiments on the domain adaption.     

\section{Conclusions}

We proposed an end-to-end framework that adapts a pre-trained 3D human pose estimation model to any target dataset by generating synthetic motions by only looking at 2D target poses. AdaptPose outperforms previous work on four public datasets by a large margin $(> 10\%)$. 
Our proposed solution can be applied to applications where limited motion data is available. Moreover, our method is able to generate synthetic human motion for other tasks such as human action recognition. The major limitation of our work is that it underperforms when there is a large body scale difference between source and train set. Although we have defined a parameter that learns to adjust the body bone lengths we observe a 10mm difference between normalized MPJPE and actual MPJPE when there is a large scale difference between source and target body scales (cross-dataset on 3DPW). Future work should address the scale ambiguity between source and target domains.
\section*{Appendix}
This Appendix provides ablations on the domain discriminator, 2D detections, and 3D discriminator. We also provide further qualitative results that compare AdaptPose against previous methods. Moreover, some failure cases of AdaptPose are visualized. 
\subsection*{Ablation: Domain Discriminator} 
\begin{figure}[h]
\begin{center}
% \fbox{\rule{0pt}{2in} \rule{.4\linewidth}{0pt}}
\includegraphics[scale=0.35]{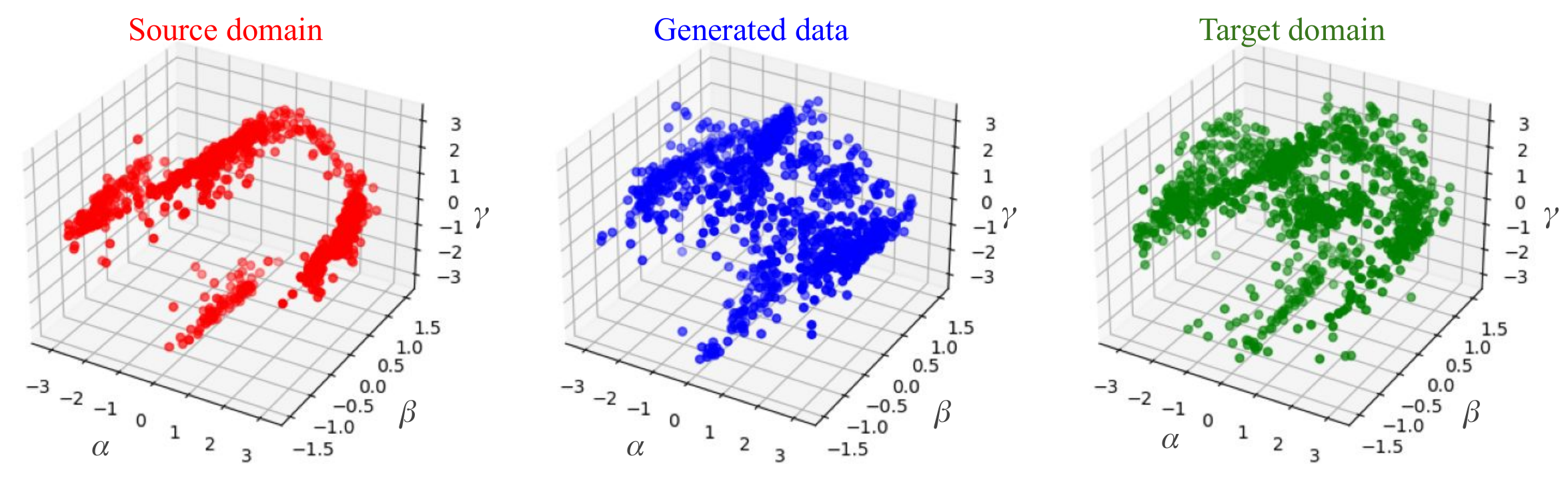}
\end{center}
  \caption{Camera viewpoints of the source (Human3.6M), target (3DHP), and generated data. The generated data follows the diversity and pattern of viewpoints of the target dataset. $\alpha$, $\beta$, and $\gamma$ are Euler-angles in radiant. The viewpoint is defined by the relative rotation matrix between the person and the camera.}
\label{fig:camera}
\end{figure}

% \HR{Name which datasets and perhaps say how their camera differes, e.g., only chest height or handheld...}
In this section, we provide further visualization of the performance of the domain discriminator. 
Figure \ref{fig:camera} shows the distribution of camera viewpoints of the source, target, and generated datasets. Human3.6M and 3DHP are the source and target datasets, respectively. Human3.6M includes four chest-view cameras, while 3DHP includes 14 cameras that cover chest-view, top-view, and bottom-view. Figure \ref{fig:camera} shows that the camera viewpoints of the target dataset are more diverse than those of the source dataset. We define the viewpoint by the relative rotation matrix between the subject and the camera. Figure \ref{fig:camera} shows that AdaptPose successfully generates camera viewpoints that follow the distribution of the target camera viewpoints.    

\subsection*{Ablation: 2D Detections}
In this section, we perform experiments on the influence of 2D detection. Using ground truth 2D for cross-dataset evaluation is the fairest comparison since most of the previous studies use the same data \cite{Gong_2021_CVPR, Guan_2021_CVPR}.
Therefore, we used ground truth 2D in our evaluations and compared our results with previous work using the same setting. 
However, ground truth 2D is not always available.
In this section, we employ AlphaPose \cite{fang2017rmpe} to obtain 2D poses of the target dataset. The model is pre-trained on MPII \cite{andriluka14cvpr} and is not fine-tuned on the target dataset.
To obtain directly comparable numbers we use the same 2D detection to evaluate 3D pose estimators of Pavllo \etal \cite{pavlakos2018learning} and Gong \etal \cite{Gong_2021_CVPR}. Table \ref{tab:2dderection} provides the cross-dataset evaluation results while using ground truth 2D and detected 2D. In this experiment, the source and target datasets are Human3.6M and 3DHP, respectively. AdaptPose outperforms other methods using both ground truth 2D and detected 2D.  
\begin{table}[h]
\small
\centering
\caption{Experiment on 2D detection. Source: Human3.6M, target: 3DHP. P2 is mean per joint position error (MPJPA) and P1 is MPJPA after Procrustes alignment of the estimated and ground truth 3D.}
\label{tab:2dderection}
\begin{tabular}{ p{2.5cm}|p{1cm}p{1cm}p{1cm}p{1cm}}
\hline
&\multicolumn{2}{c}{AlphaPose 2D}&\multicolumn{2}{|c}{GT 2D}\\
Method&P2&P1&\multicolumn{1}{|c}{P2}&P1\\
\hline
Pavllo \etal\cite{pavllo:videopose3D:2019}&86.9&127.1&\multicolumn{1}{|c}{66.5}&96.4\\
PoseAug\cite{Gong_2021_CVPR}&87.2&125.7&\multicolumn{1}{|c}{59.0}&92.6\\
Ours&\textbf{83.4}&\textbf{120.5}&\multicolumn{1}{|c}{\textbf{53.6}}&\textbf{77.2}\\
\hline
\end{tabular}
\end{table}

\subsection*{Ablation: 3D Discriminator}
\begin{figure}[h]
\begin{center}
% \fbox{\rule{0pt}{2in} \rule{.4\linewidth}{0pt}}
\includegraphics[scale=0.38]{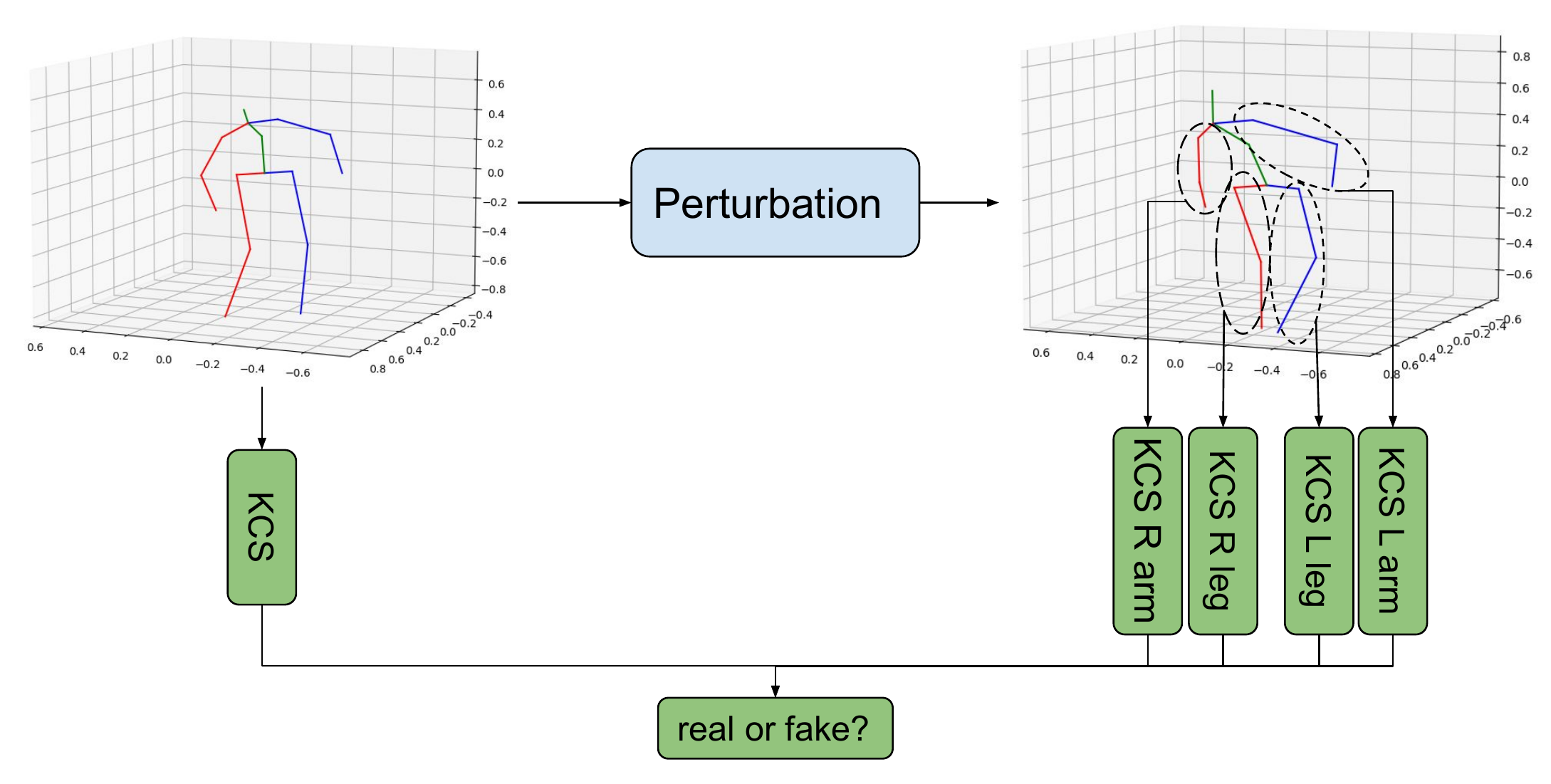}
\end{center}
  \caption{The 3D discriminator of AdaptPose.}
\label{fig:3d_disc}
\end{figure}

\begin{figure}[h]
\begin{center}
% \fbox{\rule{0pt}{2in} \rule{.4\linewidth}{0pt}}
\includegraphics[scale=0.38]{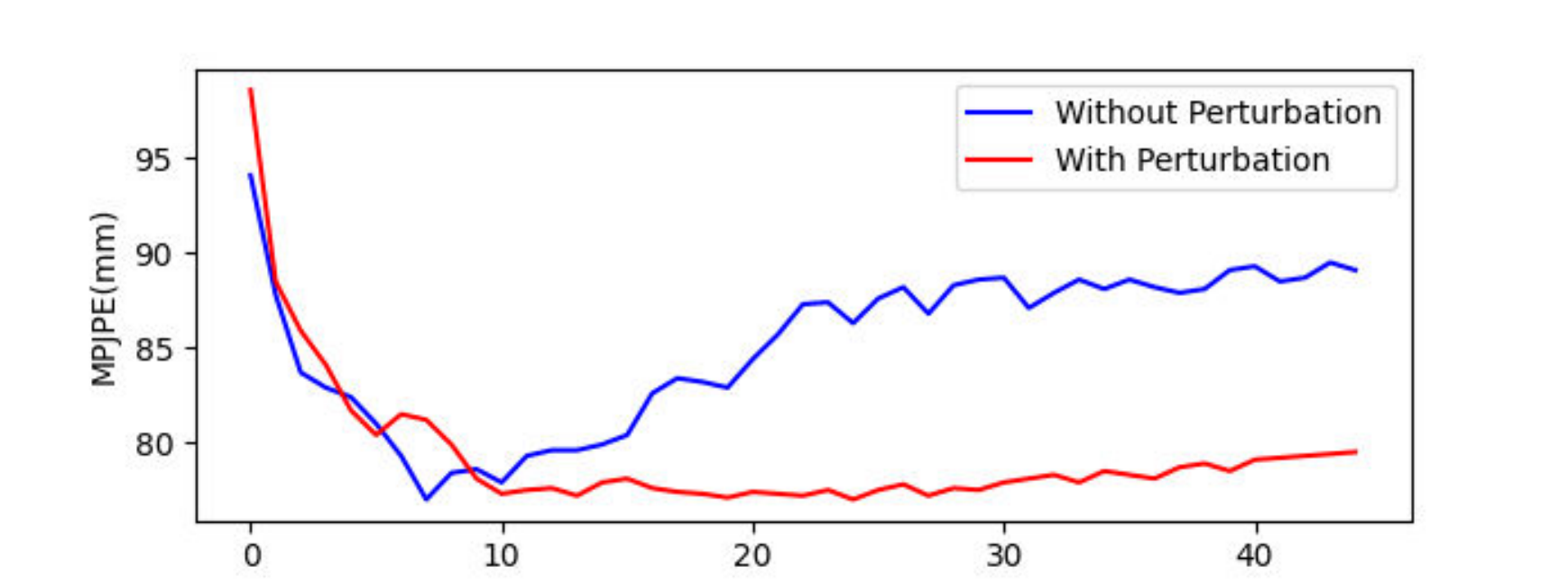}
\end{center}
  \caption{The evaluation error of AdaptPose while training with and without adding perturbation to the 3D discriminator.}
\label{fig:3d_converg}
\end{figure}

In this section we first perform an ablation on the influence of 3D discriminator. Excluding the 3D discriminator results in Figure \ref{fig:3d_disc} shows the structure of the 3D discriminator. A small perturbation $(<10\deg)$ is applied to the bone vectors of input 3D and then the perturbed version is fed to the part-wise KCS matrices of right/left arm and right/left leg. The original 3D pose is also fed to a KCS matrix. The perturbation branch enables the model to explore plausible 3D poses out of the source domain. Figure \ref{fig:3d_converg} shows the convergence curve of AdaptPose with and without perturbation of the source dataset. Without perturbation, the cross-dataset error of the model decreases to 77.2 in the first 9 epochs and then slightly increases in the following epochs. After applying the perturbation, the error decreases slower and convergence of the model is more stable. 

\begin{figure*}[h]
\begin{center}
% \fbox{\rule{0pt}{2in} \rule{.4\linewidth}{0pt}}
\includegraphics[scale=0.37]{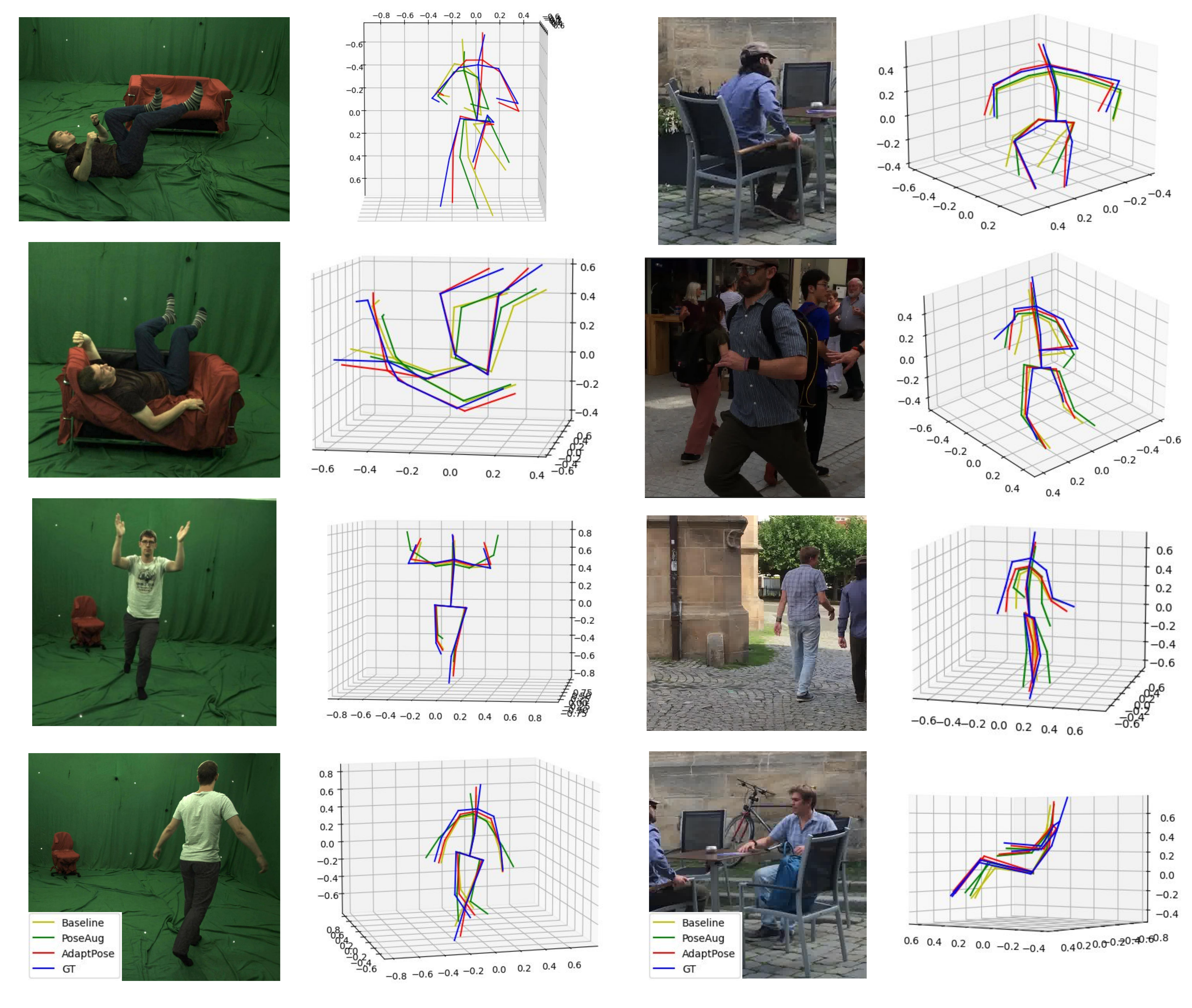}
\end{center}
  \caption{Further qualitative examples from 3DPW (right) and 3DHP (left) datasets. Yellow is Pavllo \etal \cite{pavllo:videopose3D:2019}, green is PoseAug \cite{Gong_2021_CVPR}, red is AdaptPose, and blue is the ground truth.}
\label{fig:qualitative}
\end{figure*}

\begin{figure}[h]
\begin{center}
% \fbox{\rule{0pt}{2in} \rule{.4\linewidth}{0pt}}
\includegraphics[scale=0.55]{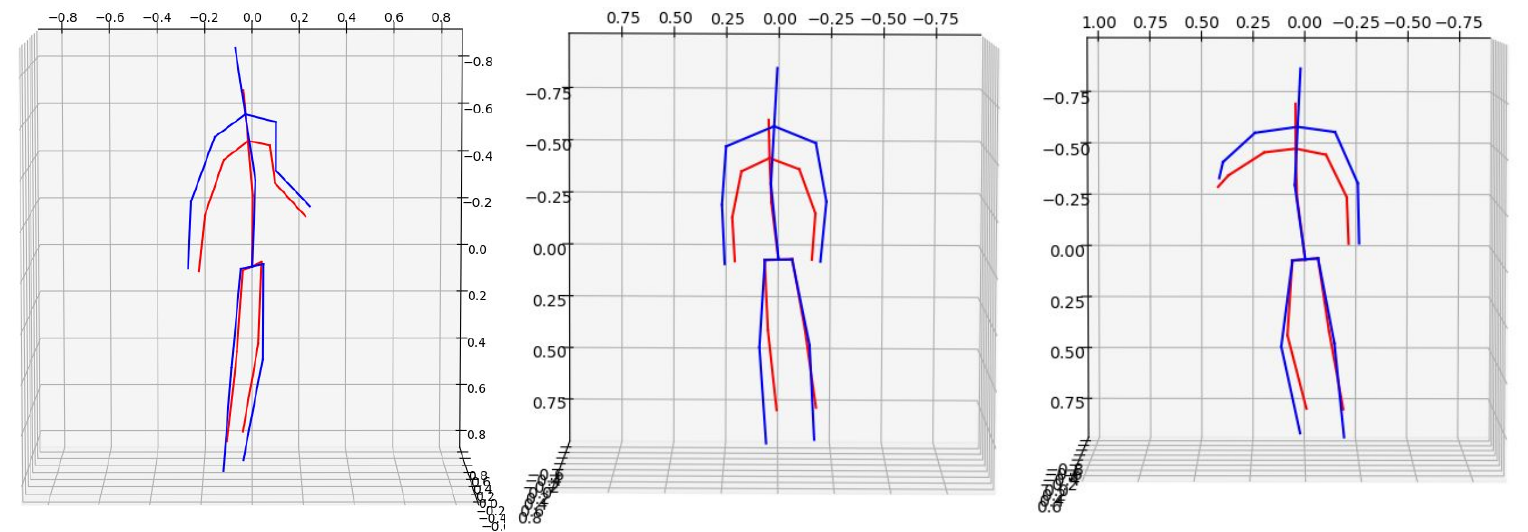}
\end{center}
  \caption{Scale error while performing cross-dataset evaluation on 3DPW dataset. Source: Human3.6M, target: 3DPW.}
\label{fig:Failure}
\end{figure}

\subsection*{Further Qualitative Results}
Figure \ref{fig:qualitative} provides further qualitative comparisons between AdaptPose, VideoPose3D\cite{pavllo:videopose3D:2019}, PoseAug\cite{Gong_2021_CVPR}, and ground truth 3D. AdaptPose significantly outperforms the previous methods. Figure \ref{fig:qualitative} shows that in the case of body occlusions, AdaptPose is more accurate than other methods. One of the main limitations of our method is scale error.  AdaptPose under-performs if there is a large difference between source and target body scales. Such scale ambiguity is inevitable when using only monocular views and no 3D supervision of the target domain is available. Figure \ref{fig:Failure} provides some examples of scale error for cross-dataset evaluation on 3DPW dataset. Code will be publicly available upon publication. 

% \section{Ablation on the Source Domain Size}

%%%%%%%%% REFERENCES
{\small
\bibliographystyle{ieee_fullname}
\bibliography{egbib}
}

\end{document}